\documentclass[10pt,twocolumn,letterpaper]{article}

\usepackage{cvpr}              

\usepackage[dvipsnames]{xcolor}

\definecolor{cvprblue}{rgb}{0.21,0.49,0.74}
\usepackage[pagebackref,breaklinks,colorlinks,citecolor=cvprblue]{hyperref}
\usepackage{pifont}
\usepackage{multirow}
\usepackage{array}

\usepackage{array}
\newcolumntype{P}[1]{>{\centering\arraybackslash}p{#1}}
\usepackage{xcolor,colortbl}

\def\paperID{9035} 
\def\confName{CVPR}
\def\confYear{2024}

\title{Guided Slot Attention for Unsupervised Video Object Segmentation}

\author{
	Minhyeok Lee \quad
	Suhwan Cho \quad
	Dogyoon Lee \quad
	Chaewon Park \quad
	Jungho Lee \quad
	Sangyoun Lee \quad
	\vspace{0.01cm}\\
	Yonsei University\\
	{\tt\small \{hydragon516,chosuhwan,nemotio,chaewon28,2015142131,syleee\}@yonsei.ac.kr}
}

\begin{document}
\maketitle
\begin{abstract}
	Unsupervised video object segmentation aims to segment the most prominent object in a video sequence. However, the existence of complex backgrounds and multiple foreground objects make this task challenging. To address this issue, we propose a guided slot attention network to reinforce spatial structural information and obtain better foreground--background separation. The foreground and background slots, which are initialized with query guidance, are iteratively refined based on interactions with template information. Furthermore, to improve slot--template interaction and effectively fuse global and local features in the target and reference frames, K-nearest neighbors filtering and a feature aggregation transformer are introduced. The proposed model achieves state-of-the-art performance on two popular datasets. Additionally, we demonstrate the robustness of the proposed model in challenging scenes through various comparative experiments. Code and models are available at \url{https://github.com/Hydragon516/GSANet}.
\end{abstract} 

\section{Introduction}
Video object segmentation (VOS) is a crucial task in computer vision, which aims to segment objects in a video sequence frame by frame. VOS is used as preprocessing for video captioning~\cite{wang2018reconstruction}, optical flow estimation~\cite{cheng2017segflow}, autonomous driving~\cite{abramov2012depth, liu2020video, maddern20171}. The VOS tasks can be divided into semi-supervised and unsupervised approaches depending on the availability of explicit target supervision. In semi-supervised VOS, the model is provided with a segmentation mask for the initial frame, and its objective is to track and segment the specified object throughout the entire video sequence. On the other hand, unsupervised VOS requires the model to find and segment the most salient objects in the video sequence without any external guidance or the initial frame mask. The unsupervised VOS is a more challenging task as it involves searching for common objects that consistently appear in the input video and effectively extracting their features.

\begin{figure}[t]
	\setlength{\belowcaptionskip}{-24pt}
	\begin{center}
		\includegraphics[width=\linewidth]{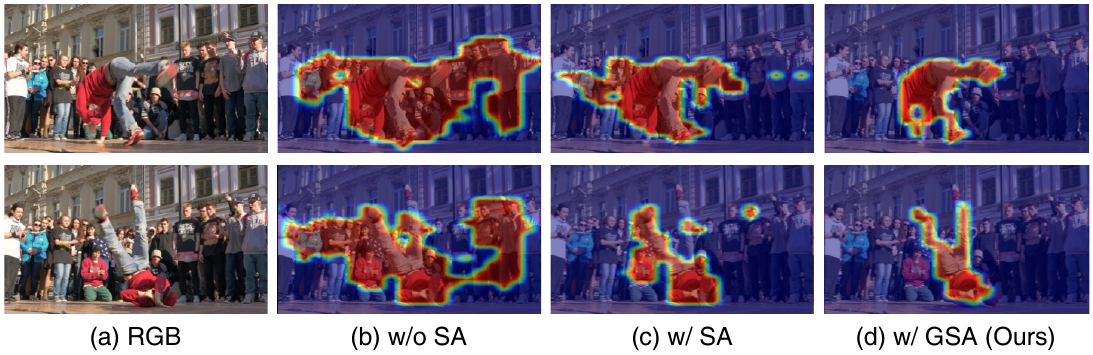}
		\vspace{-0.7cm}
		\caption{(a) Input RGB image. (b) Activation map of baseline encoder features. (c) Slot activation map of the existing slot attention method. (d) Slot activation map of the proposed guided slot attention. When guided slot attention is applied to the encoder, it surpasses the encoder's own foreground extraction ability and shows stronger performance than the previous slot attention even in complex backgrounds.}
		\label{fig:intro}
	\end{center}
\end{figure}


Due to the difficulties of unsupervised VOS, deep-learning-based unsupervised VOS models~\cite{zhou2020motion, ren2021reciprocal, ji2021full, yang2021learning, zhang2021deep, pei2022hierarchical, lee2023unsupervised, cho2023treating} have recently been in the spotlight. In particular, many approaches~\cite{zhou2020motion, ji2021full, cho2023treating, lee2023unsupervised} integrate additional motion information such as optical flow with RGB appearance information, which is motivated by the fact that the target object generally exhibits distinctive motion. These methods focus on how to properly fuse appearance and motion information. These two types of information can mutually complement each other and produce useful cues for prediction. However, they suffer from the problem that they are overly dependent on motion cues and overlook structural information of a scene such as color, texture, and shape. In cases where a scene has complex structures or quality of optical flow maps is low, those methods cannot operate reliably. 

To address these issues, we propose leveraging the slot attention mechanism originally introduced in object-centric learning. This mechanism enables the extraction of crucial spatial structural information, which is necessary for distinguishing between foreground and background, from features that are enriched with contextual information. The reason why we focus on slot attention is shared intuition between object-centric learning and unsupervised VOS that both method aims to self-learn and segment the distinguishing features of objects and backgrounds. In object centric learning, slot attention generates randomly initialized empty slots and performs iterative multi-head attention with input image features to store individual object and background information for each slot. These stored individual information of object and background in each slot provide robust foreground and background discrimination capabilities by capturing the unique characteristics and interactions of individual objects and their contexts. This intuition of discrimination can also be applied to unsupervised VOS models to increase the capabilities for discriminating the most salient object. However, the existing slot attention-based image segmentation methods~\cite{locatello2020object, zhou2022slot, yang2021self} have a significant limitation in that they work well only on synthesized images with uniform color and layout or objects that are clearly distinguished by color and shape, or simple textures such as optical flow maps, and their performance is degraded in complex real-world scenes. This limitation arises for several reasons, including: 1) randomly initialized slots are difficult to represent reasonable context in complex scenes, 2) existing simple multi-head attention operations lack robust feature discrimination capabilities, and 3) in the presence of complex backgrounds and multiple similar objects, attention to all input features can act as noise.

To tackle this limitation, we propose a novel guided slot attention network (GSA-Net) mechanism that uses guided slots, feature aggregation transformer (FAT), and K-nearest neighbors (KNN) filtering. The proposed model generates guided slots by embedding contextual information from the target frame feature of encoder, which includes coarse spatial structural information about foreground and background candidates. Our slot attention mechanism differs from existing slot attention mechanisms that employ randomly initialized empty slots as query features, and the proposed method prevents the slots from being trained in the wrong way during the initial stages of iterative multi-head attention. Additionally, providing guidance information to the slots allows the model to maintain robust context extraction ability in complex real-world scenes. Furthermore, our model extracts and aggregates global and local features from the target frame and reference frames to use as the key and value of GSA. To do this, we design FAT to create features that effectively aggregate local and global features. These features are iteratively attended with the guided slots to progressively refine the spatial information of slots by conveying rich contextual information. In this way, we complement the simple multi-head attachment of previous slot attentions to improve feature discrimination ability. In particular, the proposed slot attention employs KNN filtering to sample features close to the slot in the feature space, sequentially transmitting useful information for slot reconstruction. This stabilizes the slot refinement process in complex scenes with many objects similar to the target object, and helps generate precise reconstruction maps. In other words, our slot attention gradually samples and uses input features with high similarity to the target object, in contrast to the existing methods that use all input features simultaneously. Figure~\ref{fig:intro} demonstrates that the proposed guided slot attention maintains powerful foreground and background separation ability even in challenging scenes.

Our method was evaluated on two widely-used datasets: DAVIS-16~\cite{perazzi2016benchmark}, FBMS~\cite{ochs2013segmentation}. These datasets consist of diverse and challenging scenarios, and our proposed model achieves state-of-the-art performance on all three. Furthermore, through various ablation studies, we have demonstrated the effectiveness of our model and shown that it can achieve robust video object segmentation even in challenging sequences.

Our main contributions can be summarized as follows:
\begin{itemize}
	\item We propose a novel guided slot attention mechanism for unsupervised video object segmentation that utilizes guided slots and KNN filtering to effectively separate foreground and background spatial structural information in complex scenes.
	
	\item The proposed model generates guided slots by embedding coarse contextual information from the target frame and extracts and aggregates global and local features from the target and reference frames to refine the slots iteratively with guided slot attention.
	
	\item The proposed method achieves state-of-the-art performance on two popular datasets and demonstrates robustness in challenging sequences through various ablation studies.
\end{itemize}

\begin{figure*}[t]
	\setlength{\belowcaptionskip}{-24pt}
	\begin{center}
		\includegraphics[width=\linewidth]{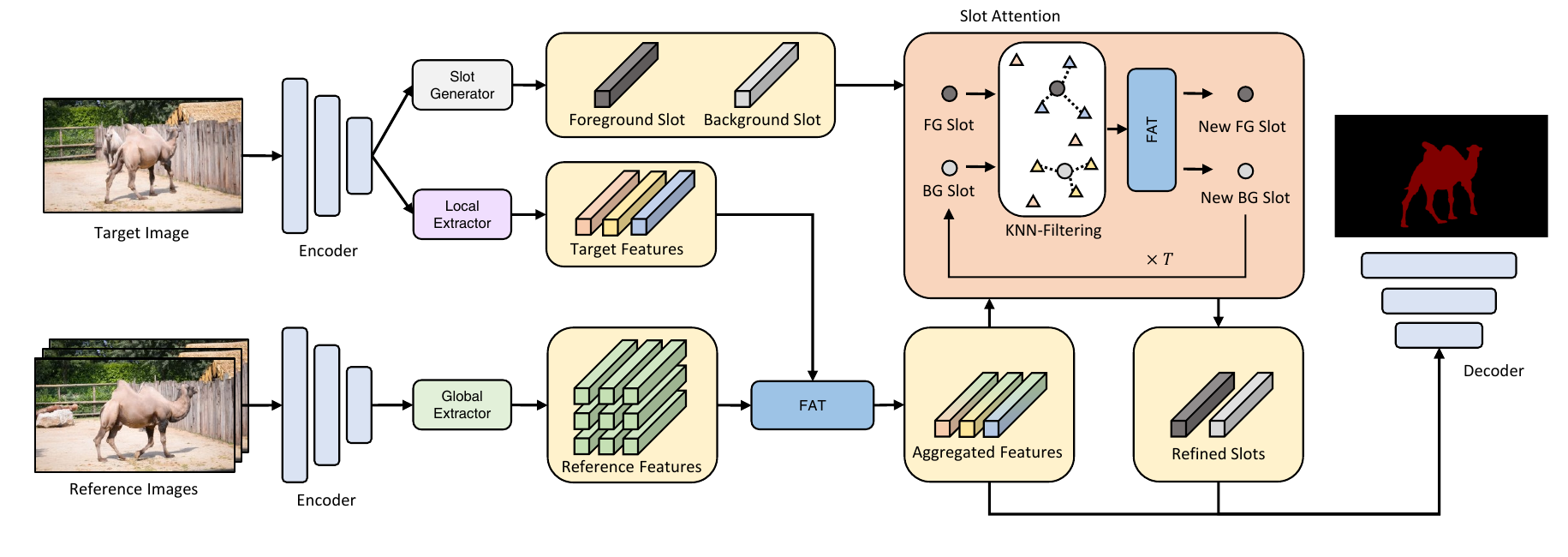}
		\vspace{-0.9cm}
		\caption{Overall structure of the proposed model. The proposed model consists of independent RGB encoder stream and optical flow encoder stream, and one decoder for mask generation. For simplicity, optical flow stream is omitted in the figure.}
		\label{fig:main}
	\end{center}
\end{figure*}

\section{Related Work}
\noindent
\textbf{Unsupervised video object segmentation.} MATNet~\cite{zhou2020motion} proposes a motion-attentive transition model for unsupervised video object segmentation. The model leverages motion information to guide the segmentation process and can segment objects. RTNet~\cite{ren2021reciprocal} presents a method based on reciprocal transformations. The propose method utilizes the consistency of object appearance and motion between consecutive frames to segment objects in a video. FSNet~\cite{ji2021full} introduces a full-duplex strategy for video object segmentation. The method uses a dual-path network to jointly model both the appearance and motion of objects in a video and can perform segmentation. AMC-Net~\cite{yang2021learning} proposes a co-attention gate that modulates the impacts of appearance and motion cues. The model learns to attend to both motion and appearance features to improve the accuracy of object segmentation. TransportNet~\cite{zhang2021deep} utilizes transport theory to model the consistency of object appearance and motion in a video. HFAN~\cite{pei2022hierarchical} introduces a hierarchical feature alignment network that aligns features from different frames at multiple scales to improve the accuracy of object segmentation. In PMN~\cite{lee2023unsupervised}, an prototype memory network is presented that utilizes a memory module to store and retrieve prototypical object representations for segmentation. The TMO~\cite{cho2023treating} treats motion as an option and can perform segmentation without relying on motion information.

\noindent
\textbf{Slot attention mechanism.}
The slot attention~\cite{locatello2020object} was first proposed for object-centric learning tasks. Object-centric learning is a type of machine learning approach where the focus is on the objects and their relationships within the context of the task. This approach has been used in various computer vision tasks, such as object detection, instance segmentation, and scene understanding.

For example, Li~\textit{et al.}~\cite{li2021scouter} propose a slot attention-based classifier for transparent and accurate classification, offering intuitive interpretation and positive or negative explanations for each category controlled by a tailored loss. Zoran~\textit{et al.}~\cite{zoran2021parts} present the model, a fully unsupervised approach for segmenting and representing objects in 3D visual scenes, which outperforms prior work through the use of a recurrent slot-attention encoder and a fixed frame-independent prior. Zhou~\textit{et al.}~\cite{zhou2022slot} present a unified end-to-end framework for video panoptic segmentation by using a video panoptic retriever to encode foreground instances and background semantics in a spatiotemporal representation called panoptic slots.

\section{Proposed Approach}
\subsection{Overall Architecture}
Figure~\ref{fig:main} shows the overall structure of the proposed GSA. The proposed model uses one target frame image and $N_R$ reference frame images as inputs. First, the slot generator generates foreground and background guidance slots from the encoded target frame image feature. These slots contain guidance information on the target foreground objects and the background. In addition, GSA extracts local features including detailed information of the target image using a local extractor and extracts global features of reference frames using a global extractor. We design an aggregation transformer to integrate this information and effectively merge the target frame features and reference frame features. Finally, the model performs slot attention using the aggregated features and guided slots. In this process, the slots are carefully adjusted by the merged features based on the KNN algorithm. As a result, the slots contain different feature information for accurate mask generation. Note that the proposed model has the same optical flow stream as the RGB image stream and this process is omitted in Figure~\ref{fig:main}. The features generated from the RGB image and optical flow are concatenated in one decoder.

\subsection{Slot Generator}
\label{sec:SG}
\begin{figure}[t]
	\setlength{\belowcaptionskip}{-24pt}
	\begin{center}
		\includegraphics[width=\linewidth]{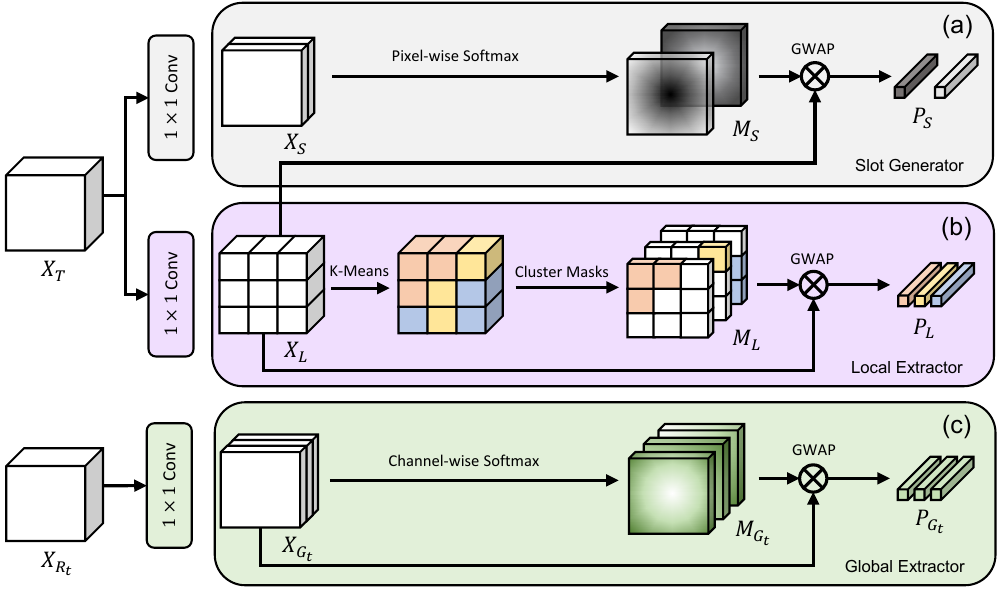}
		\vspace{-0.7cm}
		\caption{The structure of the (a) slot generator, (b) local extractor, and (c) global extractor. The slot generator creates guided slots that store important features for mask generation. The local extractor utilizes the K-means clustering algorithm to generate clustering masks at the feature level and extract local features for each region. The global extractor generates soft object regions for the scene through channel-wise softmax operations and extracts global features using these regions.}
		\label{fig:extractor}
	\end{center}
\end{figure}

Figure~\ref{fig:extractor} (a) shows the architecture of slot generator. First, the slot generator compresses the channels of the embedded target image feature $\mathbf{X_T} \in \mathbb{R} ^ {C \times H \times W}$ through a $1 \times 1$ convolutional layer to create $\mathbf{X_S} \in \mathbb{R} ^ {N_S \times H \times W}$, where $N_S$ is the number of slots. Next, slot generator applies a pixel-wise softmax operation to generate $\mathbf{M_S} \in \mathbb{R} ^ {N_S \times H \times W}$. In other words, it performs a softmax operation in the channel direction for each pixel coordinate of $\mathbf{X_S}$. Therefore, if we define the $i$-th channel of $\mathbf{X_S}$ and $\mathbf{M_S}$ as $\mathbf{X_S^i} \in \mathbb{R} ^ {1 \times H \times W}$ and $\mathbf{M_S^i} \in \mathbb{R} ^ {1 \times H \times W}$ respectively, then this process is expressed as follows:

\begin{equation}
	\mathbf{M_{S \left(x,y\right)}^i} = \frac{e^{\mathbf{X_{S \left(x,y\right)}^i}}}{\sum_{i=1}^{N_S} e^{\mathbf{X_{S \left(x,y\right)}^i}}},
\end{equation}

\noindent
where $(x, y)$ are the pixel coordinates and $i = 1, 2, ..., N_S$. Please note that $N_S=2$ because we create one slot for foreground and one slot for background. Then, we perform a global weighted average pooling (GWAP)~\cite{qiu2018global} operation between $\mathbf{M_S^i}$s and $\mathbf{X_L} \in \mathbb{R} ^ {C_L \times H \times W}$ to extract features from these feature areas, creating guided slots $\mathbf{P_S^i} \in \mathbb{R} ^ {C_L}$, where $\mathbf{X_L}$ is the target image feature embedded in the local extractor. In other words, $\mathbf{P_S^i} = \operatorname { GWAP}\left(\mathbf{X_L}, \mathbf{M_S^i}\right)$ is expressed as follows:

\begin{equation}
	\mathbf{P_S^i} = \frac{\sum_{x=1}^{H} \sum_{y=1}^{W} \left(\mathbf{M_{L \left(x,y\right)}^i} \cdot \mathbf{X_{L \left(x,y\right)}}\right)}{\sum_{x=1}^{H} \sum_{y=1}^{W} \mathbf{M_{L \left(x,y\right)}^i}}.
\end{equation}


\noindent
As a result, the slot generator creates a guided slot block $\mathbf{P_S} \in \mathbb{R} ^ {N_S \times C_L}$. Through this process, slot generator initializes the slots with useful features for final mask generation, using them as a guide. Therefore, unlike the previous slot attention method using randomly initialized slots, it is possible to create a robust and accurate mask for the object. In particular, we demonstrate in Section~\ref{ablation} that each slot after model training contains information about the foreground and background.

\begin{figure*}[t]
	\setlength{\belowcaptionskip}{-24pt}
	\begin{center}
		\includegraphics[width=0.9\linewidth]{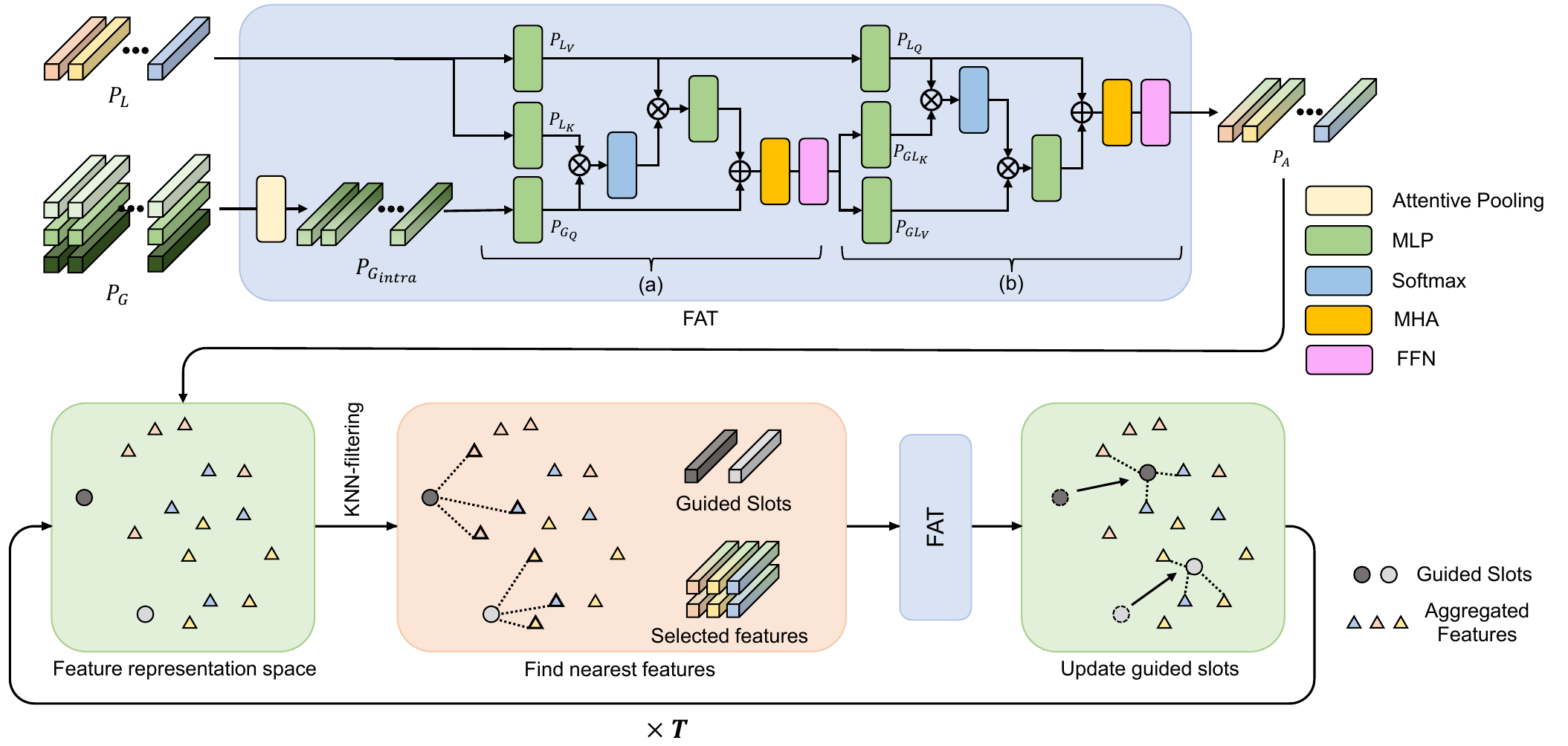}
		\vspace{-0.3cm}
		\caption{The structure of FAT and GSA. FAT uses attentive pooling to generate intra-frame features from the global features of reference frames and a transformer block to generate global to local features. GSA uses guided slots to provide initial information for foreground and background discrimination, selects the nearest features to each slot from the aggregated features using the KNN algorithm, and applies an iterative attention mechanism to update the slots. FAT and GSA aim to generate useful features for target object mask reconstruction and improve foreground and background discrimination in slot attention.}
		\label{fig:GSA}
	\end{center}
\end{figure*}

\subsection{Local \& Global Extractor}
Figure~\ref{fig:extractor} (b) and (c) show the structure of the proposed local extractor and global extractor, respectively. We use one target frame and several reference frames for training. The local extractor aims to extract detailed information of the target frame by spatially partitioning the features through feature-level k-means clustering~\cite{hartigan1979algorithm}. In addition, the global extractor generates soft object regions for each foreground object and background within the reference frames and extracts global information using these regions, taking advantage of the large amount of information.

As shown in Figure~\ref{fig:extractor} (b), the proposed local extractor performs k-means clustering on $\mathbf{X_L}$ at the pixel level to generate $D$ clustering masks $\mathbf{M_L^d} \in \mathbb{R} ^ {1 \times H \times W}$, where $d = 1, 2, ...,D$. Each mask is used to perform global weighted average pooling on $\mathbf{X_L}$ to generate $D$ local features $\mathbf{P_L^d} \in \mathbb{R} ^ {C_L} = \operatorname { GWAP}\left(\mathbf{X_L}, \mathbf{M_L^d}\right)$. As a result, the local extractor creates a local feature block $\mathbf{P_L} \in \mathbb{R} ^ {D \times C_L}$.

The structure of the global extractor, as shown in Figure~\ref{fig:extractor} (c). First, the global extractor uses the embedded feature $\mathbf{X_{G_t}} \in \mathbb{R} ^ {C_G \times H \times W}$ of the $t$-th reference frame feature $\mathbf{X_{R_t}} \in \mathbb{R} ^ {C \times H \times W}$ as input, where $t = 1, 2, ..., N_R$ and $N_R$ is the number of reference frames. Next, $\mathbf{M_{G_t}^j} \in \mathbb{R} ^ {1 \times H \times W}$ are generated from $\mathbf{X_{G_t}}$ through a channel-wise softmax operation, where $j = 1, 2, ..., C_G$. This process is expressed as follows: 

\begin{equation}
	\mathbf{M_{G_t \left(x,y\right)}^j} = \frac{e^{\mathbf{X_{G_t \left(x,y\right)}^j}}}{\sum_{x=1}^{H} \sum_{y=1}^{W} e^{\mathbf{X_{G_t \left(x,y\right)}^j}}},
\end{equation}

\noindent
and through this process, soft object regions are generated. According to~\cite{yuan2020object}, because each channel of $\mathbf{X_{G_t}}$ is generated from the convolutional kernel of a trained encoder, $\mathbf{M_{G_t}^j}$ contains approximate areas for background or foreground objects. Finally, the global extractor generates global features $\mathbf{P_{G_t}^j} \in \mathbb{R} ^ {C_G}$ through GWAP operation, similar to the slot generator and the local extractor. As a result, the global extractor creates a global feature block $\mathbf{P_G} \in \mathbb{R} ^ {N_R \times \left(C_G \times C_G\right)}$.

\subsection{Feature Aggregation Transformer}
\label{seg:FAT}
The FAT aims to generate useful features for target object mask by effectively aggregating the extracted local feature block $\mathbf{P_L}$ and global feature block $\mathbf{P_G}$. As we have extracted global features from multiple reference frames, it is important to establish the relationship between the features of the reference frames. Therefore, we use attentive pooling~\cite{hu2020randla} to consider the relationship between global features of reference frames. Through attentive pooling, intra-frame feature $\mathbf{P_{G_{intra}}} \in \mathbb{R} ^ {N_R \times C_G}$ is generated from $\mathbf{P_G}$. The part (a) in Figure~\ref{fig:GSA}, follows a standard attention structure~\cite{vaswani2017attention} $Attn\left(Q, K, V\right)$ based on queries $Q$, keys $K$, and values $V$. We use individual three multi-layer perceptron layers (MLPs) to generate $\mathbf{P_{L_K}} \in \mathbb{R} ^ {D \times C_L}$ and $\mathbf{P_{L_V}} \in \mathbb{R} ^ {D \times C_L}$, which correspond to the key and value, respectively, from $\mathbf{P_L}$ and generate $\mathbf{P_{G_Q}} \in \mathbb{R} ^ {N_R \times C_G}$, which corresponds to the query, from $\mathbf{P_{G_{intra}}}$. Finally, part (a) uses a standard transformer block composed of multi-head attention (MHA) and feed-forward network (FFN) to generate global to local feature $\mathbf{P_{GL}} \in \mathbb{R} ^ {N_R \times C_G} =\operatorname { FFN }\left(\operatorname { MHA }\left(\operatorname { Attn }\left(\mathbf{P_{G_Q}}, \mathbf{P_{L_K}}, \mathbf{P_{L_V}}\right)\right)\right)$. The part (b) in Figure~\ref{fig:GSA} has a similar configuration to the part (a). Part (b) creates a query $\mathbf{P_{L_Q}} \in \mathbb{R} ^ {D \times C_L}$ from the local feature $\mathbf{P_{L_V}}$ and creates $\mathbf{P_{GL_K}} \in \mathbb{R} ^ {N_R \times C_G}$ and $\mathbf{P_{GL_V}} \in \mathbb{R} ^ {N_R \times C_G}$, which are keys and values from $\mathbf{P_{G_L}}$, to perform attention. Also, like part (a), it creates an aggregated feature $\mathbf{P_A} \in \mathbb{R} ^ {D \times C_L}$ using MHA and FFN. As a result, $\mathbf{P_A}$ includes the features that have been integrated through local information of the target frame and global information of the reference frames.

\subsection{Guided Slot Attention}
The proposed guided slot attention is conceptually similar to previous methods as it is inspired by previous methods~\cite{locatello2020object, li2021scouter, zhou2022slot}. However, as shown in Figure~\ref{fig:GSA}, the proposed slot attention has several structural improvements. 

First, as mentioned in Section~\ref{sec:SG}, the proposed slot attention uses guided slots $\mathbf{P_S}$ generated from the slot generator. This is in contrast to previous slot attention methods that used randomly initialized empty slots. The proposed model provides initial guidance information for foreground and background discrimination by using $\mathbf{P_S}$. As a result, this leads to slots containing more accurate foreground and background features. 

Second, $N$ nearest features $\mathbf{P_{S}^{n}}$ in the feature space to each slot are selected from the aggregated features $\mathbf{P_A}$ using the K-nearest neighbors (KNN) algorithm, where $n=1, 2, ..., N$. It aims to refine the features that perform the attention operation with slots to minimize noise and stabilize learning during the attention process. On the other hand, previous slot attention computes the attention between slots and all input features. This solves the well-known problem of previous methods, where complex scenes such as many similar objects act as noise, resulting in poor performance. 

Finally, the proposed model uses an iterative attention mechanism for updating slots similar to the previous work~\cite{locatello2020object}, but we apply the FAT described in Section~\ref{seg:FAT}. The FAT performs attention between the guided slot and selected features $\mathbf{P_{S}} \in \mathbb{R} ^ {N_S \times N \times C_L}$, and updates the guided slot. $\mathbf{P_{S}}$ is applied attentive pooling to generate $\mathbf{P_{S_{intra}}} \in \mathbb{R} ^ {N_S \times C_L}$. By the attentive pooling, this process establishes the relationship between features that have the same similarity. Guided slot attention generates the final refined slot $\mathbf{P_{S_r}} \in \mathbb{R} ^ {N_S \times C_L}$ for foreground and background by repeating these three processes $T$ times: KNN filtering, attention using FAT, and slot update. This relational context information effectively integrates slots and close features through FAT, resulting in updated slots that contain more accurate foreground and background information.

\begin{table*}
	\centering 
	\caption{Quantitative evaluation on the DAVIS-16~\cite{perazzi2016benchmark} and FBMS~\cite{ochs2013segmentation}. OF and PP indicate the use of optical flow estimation models and post-processing techniques, respectively. In addition, * symbol indicates that test time augmentation is applied in the same way as the evaluation method of HFAN~\cite{pei2022hierarchical}.}
	\vspace{-0.4cm}
	\footnotesize
	\begin{tabular}{p{2cm}P{1.5cm}P{1.5cm}P{1.5cm}P{0.5cm}P{0.5cm}P{0.5cm}P{1cm}P{1cm}P{1cm}P{1cm}}
		\toprule
		\multicolumn{7}{c}{} &\multicolumn{3}{c}{DAVIS-16} &\multicolumn{1}{c}{FBMS}\\
		\cline{8-11}
		Method &Publication &backbone &Resolution &OF &PP &FPS &$\mathcal{G}_\mathcal{M}$ &$\mathcal{J}_\mathcal{M}$ &$\mathcal{F}_\mathcal{M}$ &$\mathcal{J}_\mathcal{M}$\\
		\midrule
		MATNet~\cite{zhou2020motion} &AAAI'20 &ResNet101 &473$\times$473 &\ding{51} &\ding{51} &20.0 &81.6 &82.4 &80.7 &76.1\\
		WCS-Net~\cite{zhang2020unsupervised} &ECCV'20 &EfficientNetV2 &320$\times$320 & & & 33.3 &81.5 &82.2 &80.7 &-\\
		DFNet~\cite{zhen2020learning} &ECCV'20 &DeepLabV3 &- & &\ding{51} &3.57 &82.6 &83.4 &81.8 &-\\
		F2Net~\cite{liu2021f2net} &AAAI'21 &DeepLabV3 &473$\times$473 & & &10.0 &83.7 &83.1 &84.4 &77.5\\
		RTNet~\cite{ren2021reciprocal} &CVPR'21 &ResNet101 &384$\times$672 &\ding{51} &\ding{51} &- &85.2 &85.6 &84.7 &-\\
		FSNet~\cite{ji2021full} &ICCV'21 &ResNet50 &352$\times$352 &\ding{51} &\ding{51} &12.5 &83.3 &83.4 &83.1 &-\\
		TransportNet~\cite{zhang2021deep} &ICCV'21 &ResNet101 &512$\times$512 &\ding{51} & &12.5 &84.8 &84.5 &85.0 &78.7\\
		AMC-Net~\cite{yang2021learning} &ICCV'21 &ResNet101 &384$\times$384 &\ding{51} &\ding{51} &17.5 &84.6 &84.5 &84.6 &76.5\\
		IMP~\cite{lee2022iteratively} &AAAI'22 &ResNet50 &- & & &1.79 &85.6 &84.5 &86.7 &77.5\\
		HFAN~\cite{pei2022hierarchical} &ECCV'22 &ResNet101 &512$\times$512 &\ding{51} & &19.0 &87.0 &86.6 &87.3 &-\\
		HFAN*~\cite{pei2022hierarchical} &ECCV'22 &ResNet101 &512$\times$512 &\ding{51} & &2.5 &87.6 &87.3 &87.9 &-\\
		HFAN~\cite{pei2022hierarchical} &ECCV'22 &MiT-b2 &512$\times$512 &\ding{51} & &18.4 &87.5 &86.8 &88.2 &-\\
		HFAN*~\cite{pei2022hierarchical} &ECCV'22 &MiT-b2 &512$\times$512 &\ding{51} & &2.9 &\underline{88.7} &\underline{88.0} &\underline{89.3} &-\\
		PMN~\cite{lee2023unsupervised} &WACV'23 &VGG16 &352$\times$352 &\ding{51} & &- &85.9 &85.4 &86.4 &77.7\\
		TMO~\cite{cho2023treating} &WACV'23 &ResNet101 &384$\times$384 &\ding{51} & &\textbf{43.2} &86.1 &85.6 &86.6 &79.9\\
		OAST~\cite{su2023unsupervised} &ICCV'23 &MiT-b2 &512$\times$512 &\ding{51} & &- &87.0 &86.6 &87.4 &\underline{83.0}\\
		
		\midrule
		\textbf{Ours} & &ResNet101 &512$\times$512 &\ding{51} & & \underline{41.5} & 87.7 & 87.0 & 88.4 & 79.2\\
		\textbf{Ours*} & &ResNet101 &512$\times$512 &\ding{51} & & 4.5 & 88.4 & 87.9 & 89.0 & 80.8\\
		\textbf{Ours} & &MiT-b2 &512$\times$512 &\ding{51} & & 38.2 & 88.2 & 87.4 & 87.4 & 82.3\\
		\textbf{Ours*} & &MiT-b2 &512$\times$512 &\ding{51} & & 4.1 & \textbf{88.9} & \textbf{88.3} & \textbf{89.6} &\textbf{83.1}\\
		\bottomrule
	\end{tabular}
	\label{Table:results}
\end{table*}

\subsection{Slot Decoder}
As shown in Figure~\ref{fig:main}, after guided slot attention, the model gets aggregated features $\mathbf{P_A}$ and refined slots $\mathbf{P_{S_r}}$ for foreground and background. In object-centric learning tasks, slot attention~\cite{locatello2020object} uses an autoencoder-based slot decoder for unsupervised image segmentation. However, for unsupervised video object segmentation, since we have access to ground truth masks for the target object, we design a new slot decoder based on cosine similarity of the slots. We compute the pixel-wise cosine similarity between the encoder feature and the features. The RGB stream correlation map $\mathbf{CM_{RGB}} \in \left[-1,  1\right] ^ {\left(M + N_S\right) \times H \times W}$ generated from $\mathbf{X_L}$, $\mathbf{P_A}$, and $\mathbf{P_{S_r}}$ is expressed as follows:

\begin{equation}
	\begin{aligned} 
		& \mathbf{C M_A(x, y)}=\left\{\frac{\mathbf{X_L(x, y)} \cdot \mathbf{P_A^m}}{\left\|\mathbf{X_L(x, y)}\right\|\left\|\mathbf{P_A^m}\right\|}\right\}_m, \\ 
		& \mathbf{C M_{S_r}(x, y)}=\left\{\frac{\mathbf{X_L(x, y)} \cdot \mathbf{P_{S_r}^i}}{\left\|\mathbf{X_L(x, y)}\right\|\left\|\mathbf{P_{S_r}^u}\right\|}\right\}_u, \\ 
		& \mathbf{C M_{RGB}(x, y)}=\operatorname { concat }\left(\mathbf{C M_A(x, y)}, \mathbf{C M_{S_r}(x, y)}\right),
	\end{aligned}
\end{equation}

\noindent
where $m=1, 2, ..., M$ and $u=1, 2, ..., N_S$. Also, $\operatorname { concat }\left(.\right)$ is the channel concatenation operator. Note that since we have independent encoder and slot attention streams for RGB image and optical flow map, $\mathbf{CM_{RGB}}$ for RGB and $\mathbf{CM_{FLOW}}$ for optical flow are created. Finally, The two generated $\mathbf{CM_{RGB}}$ and $\mathbf{CM_{FLOW}}$ are concatenated and passed to a CNN-based decoder to generate the final prediction mask.

\subsection{Objective Function}

We use sum of IOU loss and weighted binary cross-entropy loss as objective functions, which are often used in salient object detection tasks~\cite{wei2020f3net, zhu2022can}. This loss function helps assign more weight to the hard case pixels. The overall loss function is expressed as follows:

\begin{equation}
	\begin{aligned}
		& \mathcal{L}_{I O U}=1-\frac{\sum_{k=1}^K \min \left(\mathbf{P}_{\mathbf{k}}, \mathbf{G}_{\mathbf{k}}\right)}{\sum_{k=1}^K \max \left(\mathbf{P}_{\mathbf{k}}, \mathbf{G}_{\mathbf{k}}\right)}, \\
		& \mathcal{L}_{b c e}^w=-\sum_{k=1}^K w\left[\mathbf{G}_{\mathbf{k}} \ln \left(\mathbf{P}_{\mathbf{k}}\right)+\left(1-\mathbf{G}_{\mathbf{k}}\right) \ln \left(1-\mathbf{P}_{\mathbf{k}}\right)\right],
	\end{aligned}
\end{equation}

\noindent
where $w=\sigma\left|\mathbf{P_k}-\mathbf{G_k}\right|$ and $k$ is pixel coordinate. Also $\mathbf{G}$ and $\mathbf{P}$ are ground truth maps and prediction maps, respectively and $\mathcal{L}_{\text {total }}=\mathcal{L}_{I O U}+\mathcal{L}_{b c e}^w$.

\begin{figure*}[t]
	\setlength{\belowcaptionskip}{-24pt}
	\begin{center}
		\includegraphics[width=0.95\linewidth]{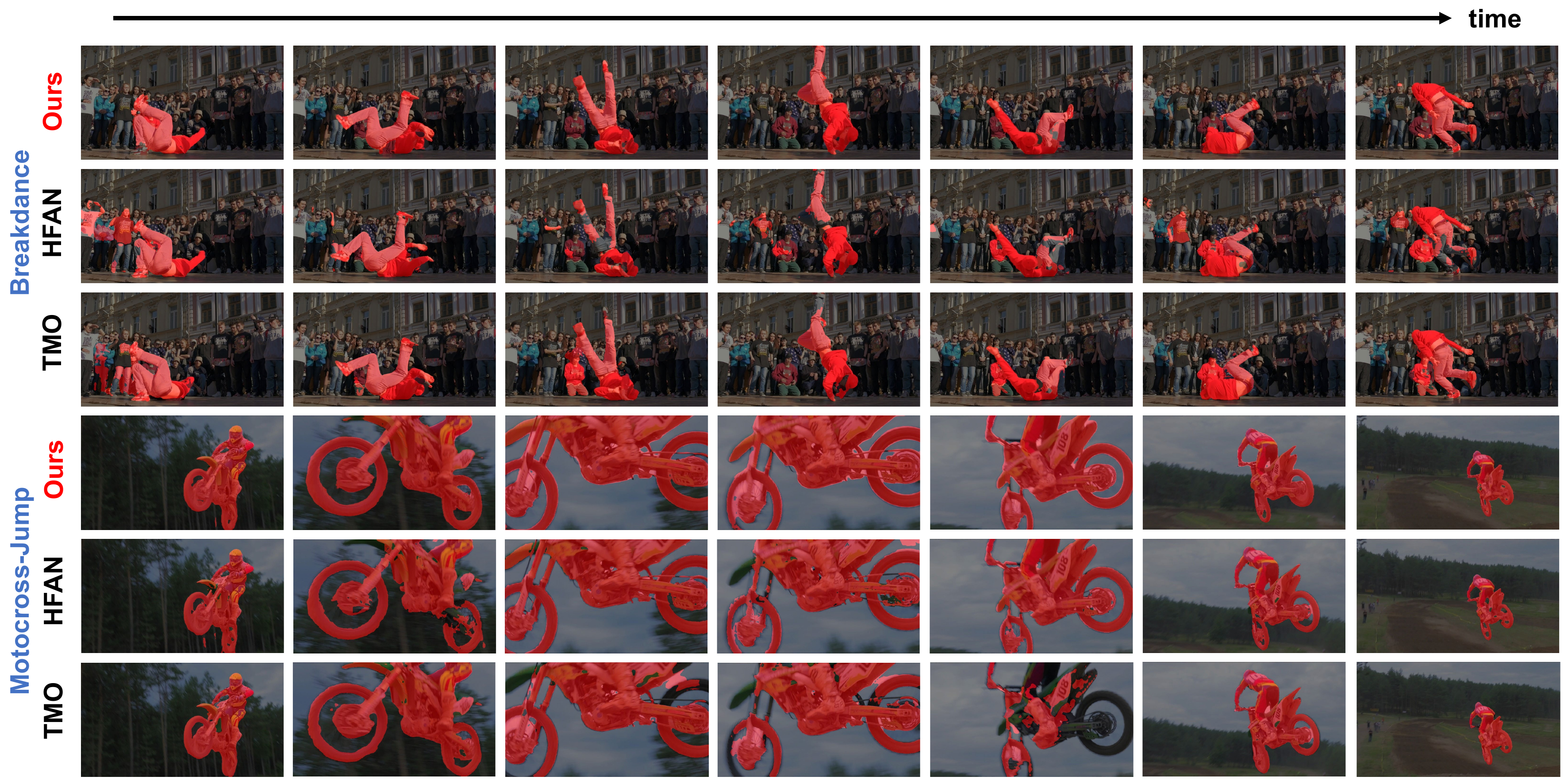}
		\vspace{-0.3cm}
		\caption{Qualitative comparison between our GSA-Net and other state-of-the-art methods.}
		\label{fig:result}
	\end{center}
\end{figure*}

\section{Experiments}
\subsection{Datasets}
In this research, we use three datasets for network training: DUTS~\cite{wang2017learning}, DAVIS-16~\cite{perazzi2016benchmark}, and YouTube-VOS~\cite{xu2018youtube}, and two datasets for network testing: DAVIS-16~\cite{perazzi2016benchmark}, FBMS~\cite{ochs2013segmentation}. The most widely used dataset is DAVIS 2016, which includes 30 training videos and 30 validation videos, and the performance of our unsupervised VOS network is primarily evaluated on the validation set of DAVIS-16~\cite{perazzi2016benchmark}. FBMS~\cite{ochs2013segmentation} is also commonly used datasets to validate the performance of VOS models.

\subsection{Evaluation Metrics}
In this study, we use three evaluation metrics to assess the performance of our method: region similarity ($\mathcal{J}$), boundary accuracy ($\mathcal{F}$), and their average ($\mathcal{G}$). The calculation of $\mathcal{J}$ and $\mathcal{F}$ is as follows:

\begin{equation}
	\mathcal{J}=\left|\frac{\mathbf{G} \cap \mathbf{P}}{\mathbf{G} \cup \mathbf{P}}\right|, \mathcal{F}=\frac{2 \times \text { Precision } \times \text { Recall }}{\text { Precision }+\text { Recall }},
\end{equation}

\noindent
where Precision $=\sum \mathbf{P} \cap \mathbf{G} / \sum \mathbf{P}$ and Recall $=\sum \mathbf{P} \cap \mathbf{G} / \sum \mathbf{G}$.

\subsection{Model Training}
Our model is trained in three steps, following the methodology of previous works~\cite{lee2023unsupervised, ji2021full, ren2021reciprocal, liu2021f2net, lu2019see}. Firstly, we utilize a well-known saliency dataset, DUTS~\cite{wang2017learning}, to pretrain the model and prevent overfitting. As the DUTS~\cite{wang2017learning} dataset does not contain optical flow maps, only the RGB encoders and decoders of the RGB stream are pretrained. Secondly, the pretrained parameters of the RGB stream are applied equally to the optical flow stream. Lastly, the entire model is fine-tuned with the training set of the DAVIS-16~\cite{perazzi2016benchmark} and YouTube-VOS~\cite{xu2018youtube} dataset. we regard them as a single object to obtain binary ground truth masks. The optical flow map required for training is generated using RAFT~\cite{teed2020raft}, a pre-trained optical flow estimation model. 

\subsection{Implementation Details}
In this paper, we set the clustering count $M$ of the local feature extractor to 64, the number of reference frames $N_R$ to 3, and the number of KNN-filtered samples $N$ in GSA to 16. In particular, we randomly sample the reference frames during the training phase and uniformly sample them during the testing phase. In addition, the number of training and testing time iterations for slot attention $T$ is set to 3, the same as in~\cite{locatello2020object}. All RGB images optical flow maps are uniformly resized to $384 \times 384$ pixels for both training and inference. The Adam optimizer~\cite{kingma2014adam} is used for network training and fine-tuning with hyperparameters $\beta_1 = 0.9$, $\beta_2 = 0.999$, and $\epsilon = 10^{-8}$. The learning rate decreases from $10^{-4}$ to $10^{-5}$ using a cosine annealing scheduler~\cite{loshchilov2016sgdr}. The total number of epochs is set to 200, with a batch size of 12. The experiments are conducted on a two NVIDIA RTX 3090 GPUs and are implemented using the PyTorch deep-learning framework.

\subsection{Results}
\noindent
\textbf{Quantitative results.} Table~\ref{Table:results} shows the quantitative results of the proposed GSA-Net. Our model is evaluated on the RenNet101~\cite{he2016deep} and MiT-b2~\cite{xie2021segformer} backbone encoders, respectively. In most conventional Unsupervised VOS methods, single-scale testing without applying test time augmentation is employed. However, for a fair comparison with HFAN~\cite{pei2022hierarchical}, we include the results of applying multi-scale testing with test time augmentation. As shown in the table, our method achieves state-of-the-art performance on both challenging datasets. In particular, compared to the HFAN~\cite{pei2022hierarchical} with $512 \times 512$ resolution, the proposed GSA-Net shows comparable performance achieving faster FPS and higher performance. In contrast to the DAVIS-16~\cite{perazzi2016benchmark}, the FBMS~\cite{ochs2013segmentation} includes both single-object and multi-object scenarios. Remarkably, even in these more complex scenarios, our proposed method, outperforms all other existing approaches with a significant margin. This result showcases the robustness of GSA-Net for handling videos with multiple objects.

\noindent
\textbf{Qualitative results.} We compared the performance of our proposed model, GSA-Net, with two state-of-the-art models, HFAN~\cite{pei2022hierarchical} and TMO~\cite{cho2023treating}, using the DAVIS-16~\cite{perazzi2016benchmark} dataset. The results, presented in Figure~\ref{fig:result}, demonstrate that GSA-Net outperforms both HFAN and TMO in various challenging video sequences. Specifically, GSA-Net shows robustness in complex background situations with many objects that are similar in appearance to the target object, as demonstrated in the \textit{Breakdance} sequence. In addition, the GSA-Net model is capable of consistent feature extraction even with extreme scale changes of the objects, as shown in the \textit{Motocross-Jump} sequence. Overall, these results suggest that GSA-Net is a promising approach for object tracking in challenging video sequences.

\begin{table}
	\centering 
	\caption{Performance with different combinations of our contributions on the DAVIS-16~\cite{perazzi2016benchmark} dataset. (a) is the baseline model, GS stands for guided slots, KNN stands for KNN filtering, and SA stands for slot attention. If GS is disabled, randomly initialized slots are used, and if FAT is disabled, the standard transformer structure of~\cite{locatello2020object} is used.}
	\vspace{-0.4cm}
	\resizebox{\columnwidth}{!}{
		\begin{tabular}{c|cccc|ccc|c}
			\hline
			\multirow{2}{*}{Index} & \multicolumn{4}{c|}{Method} & \multicolumn{3}{c|}{DAVIS-16} & \multicolumn{1}{c}{FBMS} \\ \cline{2-9} 
			& GS   & KNN   & SA   & FAT   &$\mathcal{G}_\mathcal{M}$ &$\mathcal{J}_\mathcal{M}$ &$\mathcal{F}_\mathcal{M}$ &$\mathcal{J}_\mathcal{M}$  \\ \hline
			(a) & & & & & 83.7 & 83.3 & 84.1 & 76.1\\
			(b) & & & \ding{51} & & 84.1 & 83.8 & 84.2 & 76.3\\
			(c) & \ding{51} & & \ding{51} & & 86.1 & 85.8 & 86.4 & 78.4\\
			(d) & & \ding{51} & \ding{51} & & 84.9 & 84.6 & 85.2 & 77.5\\
			(e) & \ding{51} & \ding{51} & \ding{51} & & 86.5 & 86.7 & 86.5 & 78.9\\
			(f) & \ding{51} & \ding{51} & \ding{51} & \ding{51} & \textbf{87.7} & \textbf{87.0} & \textbf{88.4} & \textbf{79.2}\\ \hline
		\end{tabular}
	}
	\label{Table:combination}
\end{table}

\begin{figure}[t]
	\setlength{\belowcaptionskip}{-24pt}
	\begin{center}
		\includegraphics[width=\linewidth]{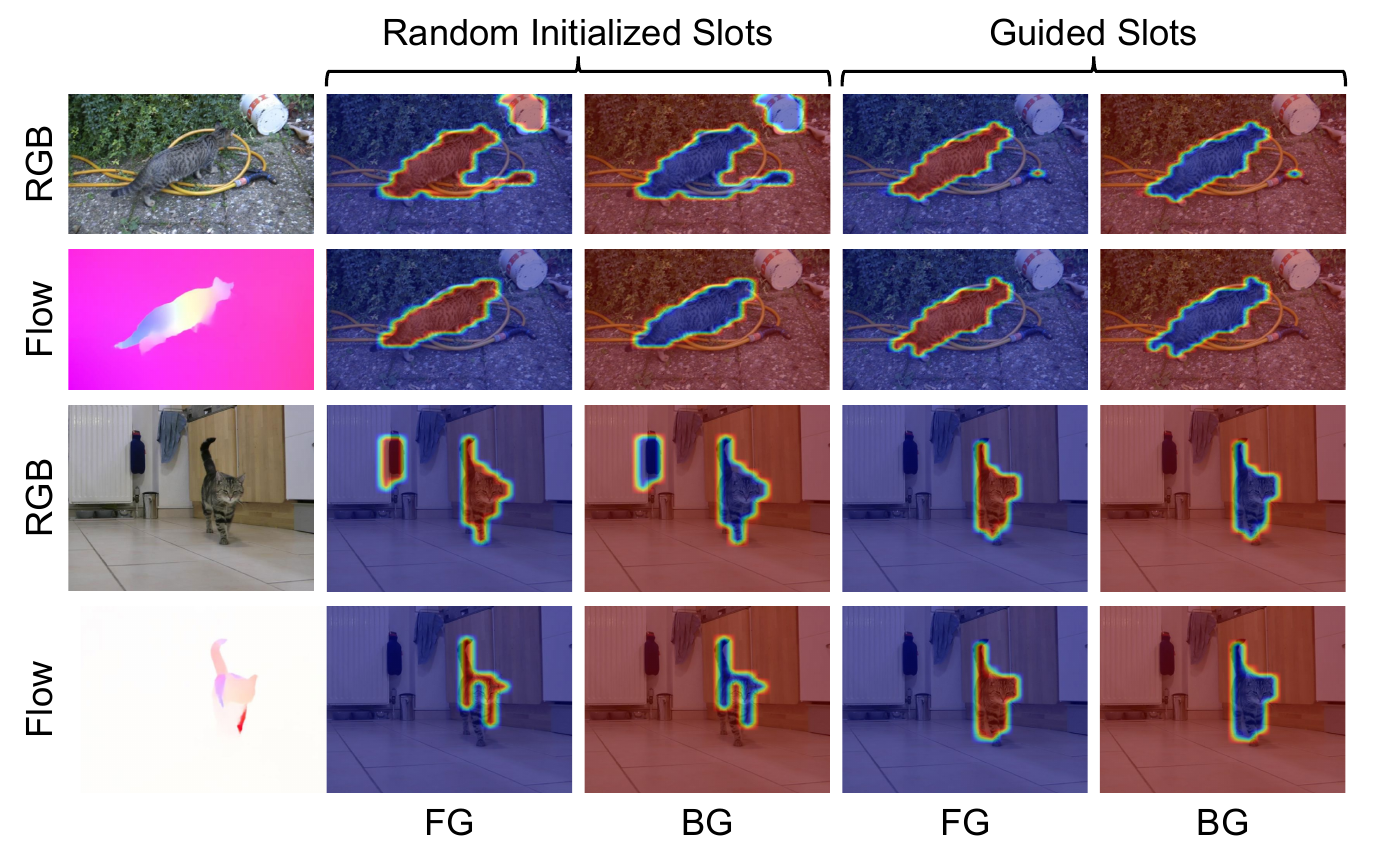}
		\vspace{-0.8cm}
		\caption{Visualization of similarity maps for final foreground (FG) and background (BG) slots depending on the use of guided slots. Evaluation is performed on both RGB images and optical flow maps.}
		\label{fig:ablation}
	\end{center}
\end{figure}

\subsection{Ablation Analysis}
\label{ablation}
This section includes various ablation experiments on the proposed model. All experiments are evaluated at the same $512 \times 512$ image resolution as the ResNet101~\cite{simonyan2014very} backbone.

\noindent
\textbf{Effect of guided slots.} Table~\ref{Table:combination} (b), (c) and Figure~\ref{fig:ablation} demonstrate the effect of the proposed guided slots. Using the slots generated by the proposed slot generator, as opposed to the existing method with randomly initialized slots, shows significant performance improvement in all evaluation metrics. Particularly, Figure~\ref{fig:ablation} shows the final refined foreground and background slot masks when using both randomly initialized slots and guided slots, which exhibits strong target object discrimination ability in complex RGB images containing multiple objects.

\noindent
\textbf{Effect of KNN filtering and FAT.} Table~\ref{Table:combination} (d), (e), and (f) demonstrate the effectiveness of the proposed KNN filtering and FAT, both of which show significant performance improvements across all evaluation metrics. In particular, FAT exhibits robust mask accuracy by effectively integrating local information from the target frame and global information from reference frames, compared to standard transformer blocks. Furthermore, KNN filtering shows a high performance improvement in FBMS with multiple target objects, demonstrating that by sampling features, it can effectively extract generalized features for multi-objects through slot attention.

\begin{figure}[t]
	\setlength{\belowcaptionskip}{-24pt}
	\begin{center}
		\includegraphics[width=\linewidth]{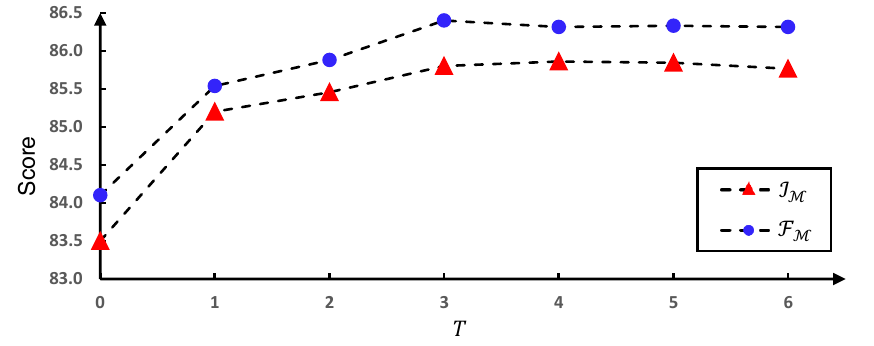}
		\vspace{-0.8cm}
		\caption{Comparison of performance characteristics with the number of iterations $T$ on the DAVIS-16~\cite{perazzi2016benchmark} dataset.}
		\vspace{0.5cm}
		\label{fig:ablation_T}
	\end{center}
\end{figure}

\begin{figure}[t]
	\setlength{\belowcaptionskip}{-24pt}
	\begin{center}
		\includegraphics[width=\linewidth]{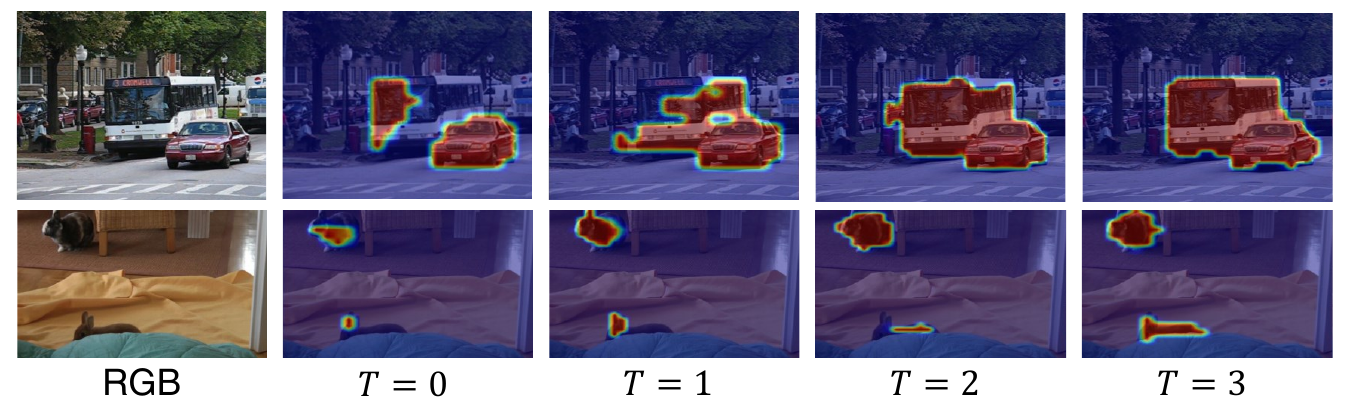}
		\vspace{-0.8cm}
		\caption{Visualization of foreground slot similarity maps with the number of iterations $T$.}
		\label{fig:ablation_T2}
	\end{center}
\end{figure}

\noindent
\textbf{Effect of number of testing time iterations.} Figure~\ref{fig:ablation_T} and~\ref{fig:ablation_T2} illustrates how the performance changes according to the iteration number $T$ of the proposed guided slot attention during the model's test stage. Proposed guided slot attention improves the quality of refined slot masks as attention mechanism is iteratively applied. Notably, the proposed method exhibits performance improvements up to three iterations, beyond which no significant changes in performance are observed. This suggests that the slots have been sufficiently refined through KNN filtering and FAT. As the number of iterations increases, the inference time of the model also increases, so $T=3$ is considered the most optimal.

\section{Conclusion}
We proposed a novel guided slot attention mechanism for unsupervised VOS. Our model generates guided slots by embedding coarse contextual information from the target frame, which allows for effective differentiation of foreground and background in complex scenes. We designed the FAT to create features that effectively aggregate local and global features. The proposed slot attention employs KNN filtering to sample features close to the slot for more accurate segmentation. Experimental results show that our method outperforms existing state-of-the-art methods.

\noindent\footnotesize\textbf{Acknowledgement.}
This work was supported by Institute of Information \& communications Technology Planning \& Evaluation (IITP) grant funded by the Korea government(MSIT) (No.2021-0-02068, Artificial Intelligence Innovation Hub) and supported by AIonFlow Research.

{
    \small
    \bibliographystyle{ieeenat_fullname}
    \bibliography{main}

\begin{thebibliography}{41}
\providecommand{\natexlab}[1]{#1}
\providecommand{\url}[1]{\texttt{#1}}
\expandafter\ifx\csname urlstyle\endcsname\relax
  \providecommand{\doi}[1]{doi: #1}\else
  \providecommand{\doi}{doi: \begingroup \urlstyle{rm}\Url}\fi

\bibitem[Abramov et~al.(2012)Abramov, Pauwels, Papon, W{\"o}rg{\"o}tter, and
  Dellen]{abramov2012depth}
Alexey Abramov, Karl Pauwels, Jeremie Papon, Florentin W{\"o}rg{\"o}tter, and
  Babette Dellen.
\newblock Depth-supported real-time video segmentation with the kinect.
\newblock In \emph{2012 IEEE workshop on the applications of computer vision
  (WACV)}, pages 457--464. IEEE, 2012.

\bibitem[Cheng et~al.(2017)Cheng, Tsai, Wang, and Yang]{cheng2017segflow}
Jingchun Cheng, Yi-Hsuan Tsai, Shengjin Wang, and Ming-Hsuan Yang.
\newblock Segflow: Joint learning for video object segmentation and optical
  flow.
\newblock In \emph{Proceedings of the IEEE international conference on computer
  vision}, pages 686--695, 2017.

\bibitem[Cho et~al.(2023)Cho, Lee, Lee, Park, Kim, and Lee]{cho2023treating}
Suhwan Cho, Minhyeok Lee, Seunghoon Lee, Chaewon Park, Donghyeong Kim, and
  Sangyoun Lee.
\newblock Treating motion as option to reduce motion dependency in unsupervised
  video object segmentation.
\newblock In \emph{Proceedings of the IEEE/CVF Winter Conference on
  Applications of Computer Vision}, pages 5140--5149, 2023.

\bibitem[Hartigan and Wong(1979)]{hartigan1979algorithm}
John~A Hartigan and Manchek~A Wong.
\newblock Algorithm as 136: A k-means clustering algorithm.
\newblock \emph{Journal of the royal statistical society. series c (applied
  statistics)}, 28\penalty0 (1):\penalty0 100--108, 1979.

\bibitem[He et~al.(2016)He, Zhang, Ren, and Sun]{he2016deep}
Kaiming He, Xiangyu Zhang, Shaoqing Ren, and Jian Sun.
\newblock Deep residual learning for image recognition.
\newblock In \emph{Proceedings of the IEEE conference on computer vision and
  pattern recognition}, pages 770--778, 2016.

\bibitem[Hu et~al.(2020)Hu, Yang, Xie, Rosa, Guo, Wang, Trigoni, and
  Markham]{hu2020randla}
Qingyong Hu, Bo Yang, Linhai Xie, Stefano Rosa, Yulan Guo, Zhihua Wang, Niki
  Trigoni, and Andrew Markham.
\newblock Randla-net: Efficient semantic segmentation of large-scale point
  clouds.
\newblock In \emph{Proceedings of the IEEE/CVF Conference on Computer Vision
  and Pattern Recognition}, pages 11108--11117, 2020.

\bibitem[Ji et~al.(2021)Ji, Fu, Wu, Fan, Shen, and Shao]{ji2021full}
Ge-Peng Ji, Keren Fu, Zhe Wu, Deng-Ping Fan, Jianbing Shen, and Ling Shao.
\newblock Full-duplex strategy for video object segmentation.
\newblock In \emph{Proceedings of the IEEE/CVF international conference on
  computer vision}, pages 4922--4933, 2021.

\bibitem[Kingma and Ba(2014)]{kingma2014adam}
Diederik~P Kingma and Jimmy Ba.
\newblock Adam: A method for stochastic optimization.
\newblock \emph{arXiv preprint arXiv:1412.6980}, 2014.

\bibitem[Lee et~al.(2023)Lee, Cho, Lee, Park, and Lee]{lee2023unsupervised}
Minhyeok Lee, Suhwan Cho, Seunghoon Lee, Chaewon Park, and Sangyoun Lee.
\newblock Unsupervised video object segmentation via prototype memory network.
\newblock In \emph{Proceedings of the IEEE/CVF Winter Conference on
  Applications of Computer Vision}, pages 5924--5934, 2023.

\bibitem[Lee et~al.(2022)Lee, Seong, and Kim]{lee2022iteratively}
Youngjo Lee, Hongje Seong, and Euntai Kim.
\newblock Iteratively selecting an easy reference frame makes unsupervised
  video object segmentation easier.
\newblock In \emph{Proceedings of the AAAI Conference on Artificial
  Intelligence}, pages 1245--1253, 2022.

\bibitem[Li et~al.(2021)Li, Wang, Verma, Nakashima, Kawasaki, and
  Nagahara]{li2021scouter}
Liangzhi Li, Bowen Wang, Manisha Verma, Yuta Nakashima, Ryo Kawasaki, and
  Hajime Nagahara.
\newblock Scouter: Slot attention-based classifier for explainable image
  recognition.
\newblock In \emph{Proceedings of the IEEE/CVF International Conference on
  Computer Vision}, pages 1046--1055, 2021.

\bibitem[Liu et~al.(2020)Liu, Cui, Chen, Zhang, and Fan]{liu2020video}
Dongfang Liu, Yiming Cui, Yingjie Chen, Jiyong Zhang, and Bin Fan.
\newblock Video object detection for autonomous driving: Motion-aid feature
  calibration.
\newblock \emph{Neurocomputing}, 409:\penalty0 1--11, 2020.

\bibitem[Liu et~al.(2021)Liu, Yu, Wang, and Zhou]{liu2021f2net}
Daizong Liu, Dongdong Yu, Changhu Wang, and Pan Zhou.
\newblock F2net: Learning to focus on the foreground for unsupervised video
  object segmentation.
\newblock In \emph{Proceedings of the AAAI Conference on Artificial
  Intelligence}, pages 2109--2117, 2021.

\bibitem[Locatello et~al.(2020)Locatello, Weissenborn, Unterthiner, Mahendran,
  Heigold, Uszkoreit, Dosovitskiy, and Kipf]{locatello2020object}
Francesco Locatello, Dirk Weissenborn, Thomas Unterthiner, Aravindh Mahendran,
  Georg Heigold, Jakob Uszkoreit, Alexey Dosovitskiy, and Thomas Kipf.
\newblock Object-centric learning with slot attention.
\newblock \emph{Advances in Neural Information Processing Systems},
  33:\penalty0 11525--11538, 2020.

\bibitem[Loshchilov and Hutter(2016)]{loshchilov2016sgdr}
Ilya Loshchilov and Frank Hutter.
\newblock Sgdr: Stochastic gradient descent with warm restarts.
\newblock \emph{arXiv preprint arXiv:1608.03983}, 2016.

\bibitem[Lu et~al.(2019)Lu, Wang, Ma, Shen, Shao, and Porikli]{lu2019see}
Xiankai Lu, Wenguan Wang, Chao Ma, Jianbing Shen, Ling Shao, and Fatih Porikli.
\newblock See more, know more: Unsupervised video object segmentation with
  co-attention siamese networks.
\newblock In \emph{Proceedings of the IEEE/CVF Conference on Computer Vision
  and Pattern Recognition}, pages 3623--3632, 2019.

\bibitem[Maddern et~al.(2017)Maddern, Pascoe, Linegar, and
  Newman]{maddern20171}
Will Maddern, Geoffrey Pascoe, Chris Linegar, and Paul Newman.
\newblock 1 year, 1000 km: The oxford robotcar dataset.
\newblock \emph{The International Journal of Robotics Research}, 36\penalty0
  (1):\penalty0 3--15, 2017.

\bibitem[Ochs et~al.(2013)Ochs, Malik, and Brox]{ochs2013segmentation}
Peter Ochs, Jitendra Malik, and Thomas Brox.
\newblock Segmentation of moving objects by long term video analysis.
\newblock \emph{IEEE transactions on pattern analysis and machine
  intelligence}, 36\penalty0 (6):\penalty0 1187--1200, 2013.

\bibitem[Pei et~al.(2022)Pei, Shen, Yao, Xie, Tang, and
  Tang]{pei2022hierarchical}
Gensheng Pei, Fumin Shen, Yazhou Yao, Guo-Sen Xie, Zhenmin Tang, and Jinhui
  Tang.
\newblock Hierarchical feature alignment network for unsupervised video object
  segmentation.
\newblock In \emph{European Conference on Computer Vision}, pages 596--613.
  Springer, 2022.

\bibitem[Perazzi et~al.(2016)Perazzi, Pont-Tuset, McWilliams, Van~Gool, Gross,
  and Sorkine-Hornung]{perazzi2016benchmark}
Federico Perazzi, Jordi Pont-Tuset, Brian McWilliams, Luc Van~Gool, Markus
  Gross, and Alexander Sorkine-Hornung.
\newblock A benchmark dataset and evaluation methodology for video object
  segmentation.
\newblock In \emph{Proceedings of the IEEE conference on computer vision and
  pattern recognition}, pages 724--732, 2016.

\bibitem[Qiu(2018)]{qiu2018global}
Suo Qiu.
\newblock Global weighted average pooling bridges pixel-level localization and
  image-level classification.
\newblock \emph{arXiv preprint arXiv:1809.08264}, 2018.

\bibitem[Ren et~al.(2021)Ren, Liu, Liu, Chen, Han, and He]{ren2021reciprocal}
Sucheng Ren, Wenxi Liu, Yongtuo Liu, Haoxin Chen, Guoqiang Han, and Shengfeng
  He.
\newblock Reciprocal transformations for unsupervised video object
  segmentation.
\newblock In \emph{Proceedings of the IEEE/CVF conference on computer vision
  and pattern recognition}, pages 15455--15464, 2021.

\bibitem[Simonyan and Zisserman(2014)]{simonyan2014very}
Karen Simonyan and Andrew Zisserman.
\newblock Very deep convolutional networks for large-scale image recognition.
\newblock \emph{arXiv preprint arXiv:1409.1556}, 2014.

\bibitem[Su et~al.(2023)Su, Song, Liu, Liu, and Liu]{su2023unsupervised}
Tiankang Su, Huihui Song, Dong Liu, Bo Liu, and Qingshan Liu.
\newblock Unsupervised video object segmentation with online adversarial
  self-tuning.
\newblock In \emph{Proceedings of the IEEE/CVF International Conference on
  Computer Vision}, pages 688--698, 2023.

\bibitem[Teed and Deng(2020)]{teed2020raft}
Zachary Teed and Jia Deng.
\newblock Raft: Recurrent all-pairs field transforms for optical flow.
\newblock In \emph{European conference on computer vision}, pages 402--419.
  Springer, 2020.

\bibitem[Vaswani et~al.(2017)Vaswani, Shazeer, Parmar, Uszkoreit, Jones, Gomez,
  Kaiser, and Polosukhin]{vaswani2017attention}
Ashish Vaswani, Noam Shazeer, Niki Parmar, Jakob Uszkoreit, Llion Jones,
  Aidan~N Gomez, {\L}ukasz Kaiser, and Illia Polosukhin.
\newblock Attention is all you need.
\newblock \emph{Advances in neural information processing systems}, 30, 2017.

\bibitem[Wang et~al.(2018)Wang, Ma, Zhang, and Liu]{wang2018reconstruction}
Bairui Wang, Lin Ma, Wei Zhang, and Wei Liu.
\newblock Reconstruction network for video captioning.
\newblock In \emph{Proceedings of the IEEE conference on computer vision and
  pattern recognition}, pages 7622--7631, 2018.

\bibitem[Wang et~al.(2017)Wang, Lu, Wang, Feng, Wang, Yin, and
  Ruan]{wang2017learning}
Lijun Wang, Huchuan Lu, Yifan Wang, Mengyang Feng, Dong Wang, Baocai Yin, and
  Xiang Ruan.
\newblock Learning to detect salient objects with image-level supervision.
\newblock In \emph{Proceedings of the IEEE conference on computer vision and
  pattern recognition}, pages 136--145, 2017.

\bibitem[Wei et~al.(2020)Wei, Wang, and Huang]{wei2020f3net}
Jun Wei, Shuhui Wang, and Qingming Huang.
\newblock F$^3$net: fusion, feedback and focus for salient object detection.
\newblock In \emph{Proceedings of the AAAI conference on artificial
  intelligence}, pages 12321--12328, 2020.

\bibitem[Xie et~al.(2021)Xie, Wang, Yu, Anandkumar, Alvarez, and
  Luo]{xie2021segformer}
Enze Xie, Wenhai Wang, Zhiding Yu, Anima Anandkumar, Jose~M Alvarez, and Ping
  Luo.
\newblock Segformer: Simple and efficient design for semantic segmentation with
  transformers.
\newblock \emph{Advances in Neural Information Processing Systems},
  34:\penalty0 12077--12090, 2021.

\bibitem[Xu et~al.(2018)Xu, Yang, Fan, Yue, Liang, Yang, and
  Huang]{xu2018youtube}
Ning Xu, Linjie Yang, Yuchen Fan, Dingcheng Yue, Yuchen Liang, Jianchao Yang,
  and Thomas Huang.
\newblock Youtube-vos: A large-scale video object segmentation benchmark.
\newblock \emph{arXiv preprint arXiv:1809.03327}, 2018.

\bibitem[Yang et~al.(2021{\natexlab{a}})Yang, Lamdouar, Lu, Zisserman, and
  Xie]{yang2021self}
Charig Yang, Hala Lamdouar, Erika Lu, Andrew Zisserman, and Weidi Xie.
\newblock Self-supervised video object segmentation by motion grouping.
\newblock In \emph{Proceedings of the IEEE/CVF International Conference on
  Computer Vision}, pages 7177--7188, 2021{\natexlab{a}}.

\bibitem[Yang et~al.(2021{\natexlab{b}})Yang, Zhang, Qi, Lu, Wang, and
  Zhang]{yang2021learning}
Shu Yang, Lu Zhang, Jinqing Qi, Huchuan Lu, Shuo Wang, and Xiaoxing Zhang.
\newblock Learning motion-appearance co-attention for zero-shot video object
  segmentation.
\newblock In \emph{Proceedings of the IEEE/CVF International Conference on
  Computer Vision}, pages 1564--1573, 2021{\natexlab{b}}.

\bibitem[Yuan et~al.(2020)Yuan, Chen, and Wang]{yuan2020object}
Yuhui Yuan, Xilin Chen, and Jingdong Wang.
\newblock Object-contextual representations for semantic segmentation.
\newblock In \emph{European conference on computer vision}, pages 173--190.
  Springer, 2020.

\bibitem[Zhang et~al.(2021)Zhang, Zhao, Liu, Liu, and Liu]{zhang2021deep}
Kaihua Zhang, Zicheng Zhao, Dong Liu, Qingshan Liu, and Bo Liu.
\newblock Deep transport network for unsupervised video object segmentation.
\newblock In \emph{Proceedings of the IEEE/CVF International Conference on
  Computer Vision}, pages 8781--8790, 2021.

\bibitem[Zhang et~al.(2020)Zhang, Zhang, Lin, M{\v{e}}ch, Lu, and
  He]{zhang2020unsupervised}
Lu Zhang, Jianming Zhang, Zhe Lin, Radom{\'\i}r M{\v{e}}ch, Huchuan Lu, and You
  He.
\newblock Unsupervised video object segmentation with joint hotspot tracking.
\newblock In \emph{Computer Vision--ECCV 2020: 16th European Conference,
  Glasgow, UK, August 23--28, 2020, Proceedings, Part XIV 16}, pages 490--506.
  Springer, 2020.

\bibitem[Zhen et~al.(2020)Zhen, Li, Zhou, Shang, Feng, Fang, and
  Quan]{zhen2020learning}
Mingmin Zhen, Shiwei Li, Lei Zhou, Jiaxiang Shang, Haoan Feng, Tian Fang, and
  Long Quan.
\newblock Learning discriminative feature with crf for unsupervised video
  object segmentation.
\newblock In \emph{Computer Vision--ECCV 2020: 16th European Conference,
  Glasgow, UK, August 23--28, 2020, Proceedings, Part XXVII 16}, pages
  445--462. Springer, 2020.

\bibitem[Zhou et~al.(2020)Zhou, Wang, Zhou, Yao, Li, and Shao]{zhou2020motion}
Tianfei Zhou, Shunzhou Wang, Yi Zhou, Yazhou Yao, Jianwu Li, and Ling Shao.
\newblock Motion-attentive transition for zero-shot video object segmentation.
\newblock In \emph{Proceedings of the AAAI Conference on Artificial
  Intelligence}, pages 13066--13073, 2020.

\bibitem[Zhou et~al.(2022)Zhou, Zhang, Lee, Sun, Li, Zhu, Yoo, Qi, and
  Han]{zhou2022slot}
Yi Zhou, Hui Zhang, Hana Lee, Shuyang Sun, Pingjun Li, Yangguang Zhu, ByungIn
  Yoo, Xiaojuan Qi, and Jae-Joon Han.
\newblock Slot-vps: Object-centric representation learning for video panoptic
  segmentation.
\newblock In \emph{Proceedings of the IEEE/CVF Conference on Computer Vision
  and Pattern Recognition}, pages 3093--3103, 2022.

\bibitem[Zhu et~al.(2022)Zhu, Li, Xie, Yan, Liang, Chen, Wei, and
  Qin]{zhu2022can}
Hongwei Zhu, Peng Li, Haoran Xie, Xuefeng Yan, Dong Liang, Dapeng Chen,
  Mingqiang Wei, and Jing Qin.
\newblock I can find you! boundary-guided separated attention network for
  camouflaged object detection.
\newblock In \emph{Proceedings of the AAAI Conference on Artificial
  Intelligence}, pages 3608--3616, 2022.

\bibitem[Zoran et~al.(2021)Zoran, Kabra, Lerchner, and Rezende]{zoran2021parts}
Daniel Zoran, Rishabh Kabra, Alexander Lerchner, and Danilo~J Rezende.
\newblock Parts: Unsupervised segmentation with slots, attention and
  independence maximization.
\newblock In \emph{Proceedings of the IEEE/CVF International Conference on
  Computer Vision}, pages 10439--10447, 2021.

\end{thebibliography}
}


\end{document}